\documentclass[a4paper]{article}

\usepackage[T1]{fontenc}
\usepackage{amssymb,amsmath}
\usepackage{algorithm2e}
\usepackage{dsfont}
\usepackage{graphicx}
\usepackage{color}
\usepackage{colordvi}

\newtheorem{thm}{Theorem}[section]
\newtheorem{prop}[thm]{Proposition}

\allowdisplaybreaks

\title{Bayesian matrix completion: prior specification}

\author{Pierre Alquier$^{(1)}$, Vincent Cottet$^{(1)}$,  \\
Nicolas Chopin$^{(1)}$, Judith Rousseau$^{(1,2)}$
\\
\small{
(1) CREST-LS ENSAE}
\\
\small{
(2) CEREMADE, Universit\'e Paris Dauphine}
}

\date{}

\begin{document}

\maketitle

\begin{abstract}
Low-rank matrix estimation from incomplete measurements recently received 
increased attention due to the emergence of several challenging applications, such as recommender systems; see in particular the famous Netflix challenge. While the
behaviour of algorithms based on nuclear norm minimization is now well
understood~\cite{Srebro1,Srebro2,Candes1,Candes2,Candes3,Gross,Rohde,Klopp,KLT}, 
an as yet unexplored avenue of research is the
behaviour of Bayesian algorithms in this context. In this paper, we briefly
review the
priors used in the Bayesian literature for matrix completion.
A standard approach is to assign an inverse gamma prior to the singular values 
of a certain singular value decomposition of the matrix of interest; this prior is conjugate. However, we show that two other types of priors (again for the singular values) may be conjugate for this model: a gamma prior, and a discrete prior. 
Conjugacy is very convenient, as it makes it possible to implement either Gibbs sampling 
or Variational Bayes. 
Interestingly enough, the maximum {\it a posteriori} for these different priors is related to the nuclear norm
minimization problems. We also compare all
these priors on simulated datasets, and on the classical MovieLens and Netflix datasets.
\end{abstract}

\section{Introduction}

We cite the introductory paper~\cite{Bennett}: ``In Oct. 2006 Netflix released
a dataset containing $10^9$ anonymous movie ratings and challenged the data
mining,
machine learning and computer science communities to develop systems that could
beat the
accuracy of its recommendation system.'' This challenge (among others) generated
a lot of excitement in the statistical community, and an increasing interest in the matrix
completion problem. Seeing users as rows and movies as columns, the problem reduces to recovering a full matrix
based on only a few of its entries. While, in general, this task is impossible,
it becomes
feasible when the matrix has low rank. In the Netflix problem,
this amounts to assume (reasonably) the existence of a small number of typical
patterns among users, {\it eg}, those who like a particular type of movie. 
Note however that this recommendation system problem was studied since
the 90s
through collaborative filtering algorithms. An example is given
by~\cite{herlocker}
on the open dataset MovieLens (available online {\it
http://grouplens.org/datasets/movielens/}).
The first attempt to perform recommendation through low-rank
matrix completion
algorithms is due to~\cite{Srebro1,Srebro2}.

The methods used in this model usually rely on minimization of a measure of the
fit to the
observations penalized by the rank or the nuclear norm of the matrix (the
nuclear norm
is actually to be preferred as it leads to computationally feasible methods).
A ground breaking result came from Cand\`es and Tao~\cite{Candes2} and Cand\`es
and
Recht~\cite{Candes3} when they exhibited conditions ensuring that the recovery
of the
matrix from a few experiments can be perfect. This result was extended to the
context
of noisy observations (in this case, the recovery of the matrix is not exact)
in~\cite{Candes1,Gross} and efficient algorithms are proposed for example
in~\cite{Recht-Re}.

A recent series of paper study a more general problem called trace regression,
including
matrix completion, as well as other popular models (linear regression,
reduced rank regression
and multi-task learning) as special cases~\cite{Rohde,Klopp,KLT}. These papers
propose
nuclear-norm penalized estimators, derive the reconstruction error of this
method and also
prove that this error is minimax-optimal: basically, the average quadratic error
on the entries of an $m_1\times m_2$ matrix with rank $r$ from the observation
of $n$ entries
cannot be better than $(m_1 \vee m_2) r /n$, where we use the notation
$a\vee b = \max(a,b)$ and $a\wedge b=\min(a,b)$ for any real numbers $a,b$.

Bayesian estimation  is also possible in this context. Based on priors defined for
reduced rank regression~\cite{Geweke} and multi-task learning~\cite{Yu1}, several
authors proposed various Bayesian estimators for the matrix completion
problem~\cite{Lim,Salakhutdinov2,Lawrence,Yu2,Paisley,Zhou,Babacan}. The
computational implementation of these estimators 
relies on either Gibbs sampling~\cite{Salakhutdinov2} or Variational Bayes (VB)
methods~\cite{Lim},
and these algorithms are fast enough to deal with such large datasets as Netflix
or MovieLens. However, and contrary to penalized minimization methods, there has been
little research on the theoretical properties of these Bayesian estimators; in
particular on their consistency.  In fact, our simulations suggest that the consistency
will depend on the tuning of the hyperparameters in the prior.

Our contribution in this paper is twofold:
\begin{enumerate}
 \item We study various families of prior distributions on the parameters which lead
 to tractable posterior distributions and thus to feasible algorithms, be them based
 on Gibbs sampling or VB approaches. 
\item We compare these different priors on simulated and real datasets.
\end{enumerate}

The paper is organized as follows. In Section~\ref{section_priors}
we introduce the notations, review the conjugate priors in the litterature
and introduce the gamma and discrete priors.
The link between the maximum {\it a posteriori} (MAP) and the penalized
minimization
problems of~\cite{Srebro1,Srebro2,Candes1,Candes2,Candes3,Gross,
Rohde,Klopp,KLT}
is discussed in Section~\ref{section_MAP}. A simulation study is provided in
Section~\ref{section_simulations} in order to illustrate the strengths and
weaknesses
of each prior. The estimator that performs the best on the simulated datasets
is tested on the MovieLens and Netflix dataset
in Section~\ref{section_realdata}. Finally, some proofs 
are postponed to Section~\ref{section_proofs}.

\section{Notations and priors}
\label{section_priors}

\subsection{Notations}

Given a matrix $M$, $M_{i,\cdot}$ will denote the $i$-th row
of $M$ and $M_{\cdot,h}$ will denote the $h$-th column of $M$. Given a vector
$x=(x_1,\dots,x_d)$, $\mathbf{diag}(x)$ will denote the $d\times d$ matrix
$$ \mathbf{diag}(x)
=\left(
\begin{array}{c c c}
 x_1 & \dots & 0 \\
 \vdots & \ddots & \vdots \\
 0 & \dots & x_d
\end{array}
\right).
$$

Let  $m_1$, $m_2$ and $n$  denote respectively the dimensions of the matrix $\theta$
and the number of observed entries, and define $\Theta$ as the set of $m_1\times m_2$
matrices with real coefficients $\theta=(\theta_{i,j})_{1\leq i
\leq m_1,1\leq j
\leq m_2}$. We fix an integer $K\leq m_1\wedge m_2$ and
we assume than the unknown matrix $\theta^0$  to be estimated may be written as
$$ \theta^0 = M^0 (N^0)^T $$
where $M^0$ is $m_1\times K$ and $N_0$ is $m_2 \times K$ and, for some integer
$r\in\{1,\dots,K\}$, $$ M^0 = \left(M^0_{\cdot,1} |\dots | M^0_{\cdot,r} |0 |
\dots | 0 \right)
\text{ and } N^0 = \left(N^0_{\cdot,1} |\dots | N^0_{\cdot,r} |0 | \dots | 0
\right) $$
(thus $M^0_{\cdot,h}$ and $N^0_{\cdot,h}$ are null when $h>r$).
Note that when $K=m_1\wedge m_2$, any matrix can be decomposed in such a way,
with $r={\rm rank}(\theta^0)$.
Here $r$ is unknown together with $M^0$, $N^0$ and $\theta^0$.

The observations are supposed to be distributed according to the following model: 
first, $n$ pairs $(i_k,j_k)$, $1\leq k\leq n$, are drawn uniformly in
 $\{1,\dots,m_1\}\times\{1,\dots,m_2\}$, and observed with noise: 
\begin{equation} \label{model}
 Y_{k} = \theta^0_{i_k,j_k} + \varepsilon_{k} 
 \end{equation}
for any $k\in\{1,\dots,n\}$, where the $\varepsilon_{k}$ are centered independent random
variables drawn from a common distribution, for which the only assumption is that
it is sub-Gaussian with a known parameter $\sigma^2$. By this, we mean that
$\log \mathbb{E}\exp(t\varepsilon_k)
  \leq t^2 \sigma^2 / 2$. We denote by $\mathbb P $ the distribution
  induced by \eqref{model}. We summarize the observation as
  $Y=(i_k,j_k,Y_k)_{k\in\{1,\dots,n\}}$.

Note that the sub-Gaussian assumption is rather general: it
encompasses the classical Gaussian noise $\varepsilon_{k}\sim\mathcal{N}(0,s^2)$ with $s\leq
\sigma$ as
well as bounded noise. For a dataset like Netflix, 
bounded noise makes more sense as we know that the observed ratings are between 1 and 5.

We end this subsection with two remarks. First, it must be kept in mind that in
most
applications, it does not make sense to assume that $n\rightarrow\infty$ with
$m_1$ and
$m_2$ fixed - for $n$ large enough, we observe all the entries in the matrix, so
this is
no longer a matrix completion problem. In the Netflix prize, $m_1=480,189$
users,
$m_2=17,770$ movies and $n=100,480,507$ ratings over $m_1 m_2= 8,532,958,530$
entries
in the matrix, which means that less than $1.2\%$ of the entries of the matrix
are
observed. So, it might be more sensible to assume that
$m_1,m_2\rightarrow\infty$ with $n$
(and even $r$).
All theoretical results on non Bayesian methods take this into account.

To define a Bayesian procedure we must specify a likelihood, i.e. the distribution of the
noise $\varepsilon_k$. In practice, this distribution is often unknown. Following
the PAC-Bayesian approach, we use a Gaussian likelihood as a proxy for the true likelihood.
Thus, for a given prior $\pi({\rm d}\theta )$, we consider the following posterior
distribution: 
\begin{equation}
\label{posterior}
\rho_{\lambda}(d\theta) \propto
\exp\left[ - \frac{\lambda}{n} \sum_{k=1}^n(Y_{k} - \theta_{i_k,j_k})^2  \right]
\pi({\rm d}\theta )
. 
\end{equation}
Taking $\lambda=n/{2s^2}$ leads to the usual posterior when
the noise is $\mathcal{N}(0,s^2)$. However,
according to the PAC-Bayesian
theory~\cite{Catoni2003,Catoni2004,Catoni2007,A,Suzuki,AG,AB},
and more generally to works on aggregation with exponential
weights~\cite{DalTsy,Salmon,Rigollet}, one
might want to consider  a smaller $\lambda$, which usually allow to
prove the consistency of the method for any sub-Gaussian noises~\cite{Catoni2003}.

Our estimator of $\theta$ will be the posterior mean:
$$ \hat{\theta}_{\lambda} = \int_{\Theta} \theta \rho_{\lambda}(d\theta).
$$

\subsection{Priors} \label{subsec:prior}

All the aforementioned papers on Bayesian matrix completion assign 
conditional Gaussian priors to $M \in \mathbb R^{m_1\times K}$
 and $N \in \mathbb R^{m_2\times K}$, with 
$$ \theta = M N^T = \sum_{h=1}^{K} M_{\cdot,h} N_{\cdot,h}^T. $$
More precisely, given some vector $\gamma=(\gamma_1,\dots,\gamma_K)$ with
positive entries
 the columns  $ M_{\cdot,h} \sim \mathcal{N}(0,\gamma_h I_{m_1}) $ and
$ N_{\cdot,h} \sim \mathcal{N}(0,\gamma_h I_{m_2}) $ independently, where $I_m$ denotes the
$m\times m$ identity matrix.
 
Note that we usually expect that ${\rm rank}(\theta)=K$. However, if the
probability
distribution on $M$ and $N$ ensures that most of the $M_{\cdot,h}$ and
$N_{\cdot,h}$ are close to $0$, then $\theta$ will be close to a lower rank matrix.

On the other hand, Bayesian matrix completion methods usually differ in the choice of the prior distribution on $\gamma$ and on $r$.  
\begin{enumerate}
 \item As summarized in the survey paper~\cite{Geweke}, in econometrics, for the
estimation of
simultaneous equation models, and then later for the reduced rank regression
model,
it is reasonable to assume that the rank $r={\rm rank}(\theta^0)$ is known. In
this case,
it makes sense to fix $K=r$ and to consider a constant $\gamma$,
e.g. $\gamma=(1,\dots,1)$. We  denote
by
$\delta_1$ this prior (the Dirac mass at $1$).

\item For large scale matrix completion problems, it does not make sense to
assume
 that the rank $r$ is known. In this case, a reasonable approach is to set $K$
to a ``large'' value,
 so that $K\geq r$, and to consider $\gamma=(\gamma_1,\dots,\gamma_K)$ itself as
random.
 Using a prior that would enforce many of the $\gamma_h$ to be close to $0$
would lead
the prior to give more weight on (approximately) low-rank matrices. This is the
approach followed in~\cite{Lim,Salakhutdinov2,Lawrence,Yu2,Paisley,Zhou,Babacan}. Up to minor variants,
 all these authors propose to consider the $\gamma_h$ as iid inverse gamma with
parameters
 $a$ and $b$, denoted by $\Gamma^{-1}(a,b)$
as it leads to conjugate marginal distributions for $\gamma_h$. (Among the possible
variants,
 some authors consider $ M_{\cdot,h} \sim \mathcal{N}(0,\gamma_h I_{m_1}) $ and
$ N_{\cdot,h} \sim \mathcal{N}(0,\gamma_h' I_{m_2})$ for different vectors
$\gamma_h$ and $\gamma_h'$ but this does not seem to give a better estimation of the unknown rank of the matrix; simulations tend to confirm this intuition). In order to ensure that $K\geq r$ holds, it seems
natural to take $K$ as large as possible, $K=m_1\wedge m_2$, but
this may computationally prohibitive if $K$ is large. Note that it is proven
in~\cite{Alquier} that this prior leads to consistency in a different but
related model (reduced rank regression). We expect it to lead to consistency
in the matrix completion problem as well.

\item Based on a similar idea in the case of the Bayesian
LASSO~\cite{Park-Casella}, one may assign instead to the $\gamma_h$'s independent gamma
priors $\Gamma((m_1+m_2+1)/2,\beta^2 /2)$ for some $\beta>0$. 
This leads to simple close-form expressions for the conditional distribution
of $\gamma|M,N$ as discussed in the following Section. 

\item Finally, the following  prior with finite support has not  yet been considered in the literature on matrix completion
 $$ \gamma_h \sim (1-p) \delta_{\varepsilon} + p \delta_C, $$
 where $\varepsilon$ is a small positive constant, $p \in (0,1) $ and $C >> \varepsilon$. This prior is similar in spirit to the spike and slab prior for variable selection \cite{mitchell1988bayesian,GeorgeMcCulloch}.

\end{enumerate}

\subsection{Conjugacy, Gibbs sampling}

The four types of prior distributions discussed in the previous section
lead to the same conditional posterior distributions for the rows
$M_{i,\cdot}$ and $N_{j,\cdot}$ (conditional on $\gamma=(\gamma_1,\dots,\gamma_K)$
and on the data), which we now describe.

For $1\leq i\leq m_1$, let $\mathcal{V}_{i,N,\gamma}$ be the $K\times K$ matrix given by
$$ \mathcal{V}_{i,N,\gamma}^{-1} = \mathbf{diag}(\gamma)^{-1}
+  \frac{2\lambda}{n} \sum_{k:i_k=i} N_{j_k,\cdot} N_{j_k,\cdot}^T $$
and $\mathbf{m}_{i,N,\gamma}$ the $1 \times K$ vector given by
$$
\mathbf{m}_{i,N,\gamma}^T = \frac{2\lambda}{n} \mathcal{V}_{i,N,\gamma}
\sum_{k:i_k=i} Y_{i_k,j_k} N_{j_k,\cdot}^T.
$$
Then, given $N$, $\gamma$ and $Y$, the rows of $M$ are independent and
\begin{equation}\label{eq:cond_post_Mh}
M_{i,\cdot}^T|N,\gamma,Y
\sim\mathcal{N}\left(\mathbf{m}_{i,N,\gamma}^T,\mathcal{V}_{i,N,\gamma}
 \right).
\end{equation}
Similarly, for $1\leq j \leq m_2$,
let $\mathcal{W}_{j,M,\gamma}$ be the $K\times K$ matrix given by
$$ \mathcal{W}_{j,M,\gamma}^{-1} = \mathbf{diag}(\gamma)^{-1}
+ \frac{2\lambda}{n} \sum_{k:j_k=j} M_{i_k,\cdot} M_{i_k,\cdot}^T $$
and $\mathbf{n}_{j,M,\gamma}$ the $1 \times K$ vector given by
$$
\mathbf{n}_{j,M,\gamma}^T = \frac{2\lambda}{n} \mathcal{W}_{j,M,\gamma}
\sum_{k:j_k=j} Y_{i_k,j_k} M_{i_k,\cdot}^T.
$$
Then, given $M$, $\gamma$ and $Y$, the rows of $N$ are independent and
\begin{equation}\label{eq:cond_post_Nh}
N_{j,\cdot}^T|M,\gamma,Y
\sim\mathcal{N}\left(\mathbf{n}_{j,M,\gamma}^T,\mathcal{W}_{j,M,\gamma}
 \right).
\end{equation}

On the other hand, the priors discussed in the previous section
generate different conditional posterior distributions for $\gamma$ given 
$M$, $N$ and the data, which are summarized in Table~\ref{tab:post}
\begin{table}
\caption{Conditional posterior distribution of $\gamma$ given $M$ and $N$ for
different priors.}
\label{tab:post}
\begin{center}

\begin{tabular}{|c|c|}
\hline
 Prior on $\gamma_h$ & Conditional posterior \\
\hline
 $\delta_1$ & $\delta_1 $ \\
\hline
 $\Gamma^{-1}(a,b)$ & $\Gamma^{-1}(\hat{a}_h,\hat{b}_h)$ \\
 \hline
 $\Gamma((m_1+m_2+1)/2,\beta^2 /2)$ &
$\mathcal{IG}(\hat{\mu}_h,\hat{\lambda}_h)$ \\
 \hline
 $ (1-p) \delta_{\varepsilon} + p \delta_C $ &
 $ (1-\hat{p}_h) \delta_{\varepsilon} + \hat{p}_h \delta_C $
 \\
 \hline
\end{tabular}
\end{center}
\end{table}
where $\mathcal{IG}(\mu,\lambda)$ denotes the inverse Gaussian distribution with
parameters $(\mu,\lambda)$ and
\begin{align*}
\hat{a}_h & = a + \frac{m_1+m_2}{2}, &
\hat{b}_h & = b + \frac{\|M_{\cdot,h} \|^2 + \|N_{\cdot,h} \|^2}{2},
\\
\hat{\mu}_h & = \frac{\beta}{\sqrt{\|M_{\cdot,h} \|^2 + \|N_{\cdot,h} \|^2}}, &
\hat{\lambda}_h & = \beta^2,
\end{align*}
and finally,
$$ \hat{p}_h = \frac{\pi_h}{\pi_h+\pi_h'} $$
with
\begin{align*}
 \pi_h & =  \frac{p}{C^{(m_1+m_2)/2}}
                  \exp\left(- \frac{\|M_{\cdot,h} \|^2 + \|N_{\cdot,h} \|^2}
                   {2C} \right)
                   \\
 \pi_h' & = \frac{1-p}{\varepsilon^{(m_1+m_2)/2}}
                    \exp\left(- \frac{\|M_{\cdot,h} \|^2 + \|N_{\cdot,h} \|^2}
                   {2\varepsilon} \right).
\end{align*}

We skip the calculations that lead to these expressions, as they are a bit tedious
and follow from first principles. The surprising result in this array is the simple expression obtained for the less common gamma prior (third row). We shall see in our simulations that this gamma
prior actually leads to better performance than the more standard inverse gamma
prior that has been used in most papers on Bayesian matrix completion. 

Of course, the main motivation for deriving these conditional posterior distributions
is to be able to implement Gibbs sampling to simulate from the joint posterior of $M$, $N$, and $\gamma$ (and therefore $\theta$). The corresponding Gibbs sampler may be summarised as: 

\begin{enumerate}
\item Simulate each row $M_{i,\cdot}$ from \eqref{eq:cond_post_Mh}.
\item Simulate each row $N_{j,\cdot}$ from \eqref{eq:cond_post_Nh}.
\item Simulate $\gamma$ from the appropriate distribution from Table \ref{tab:post}. 
\end{enumerate}

Given the typical size of matrix completion problems, it is essential to be able
to implement a Gibbs sampler that updates jointly large blocks of random variables, 
as any other type of MCMC sampler (such as Metropolis-Hastings) would be likely
to show very poor performance on such high-dimensional problems. 

\subsection{Variational Bayes}

Using (conditionally) conjugate priors makes it also possible to quickly obtain a VB
(Variational Bayes) approximation of the posterior, which is convenient when
Gibbs sampling is too expensive, either because of bad mixing, or a high cost
per iteration (large datasets), or both. 

VB amounts to compute iteratively the  optimal approximation of $\rho_{\lambda}(M,N,\gamma)$ among a certain class of distributions; in our case, the class of factorized distributions $q(M,N,\gamma)=q(M)q(N)q(\gamma)$. The optimality criterion is 
the K\"ullback-Leibler divergence between $q(M,N,\gamma)$ and
$\rho_{\lambda}(M,N,\gamma)$, $\mathcal{K}(q,\rho_{\lambda})$. The algorithm
works iteratively, by updating each factor $q(M)$, $q(N)$, and $q(\gamma)$, in turn. 

Here again, we skip the tedious but elementary calculations. Note that, in the simulation
section, we use this algorithm only with the inverse-gamma prior, so we only describe
this version of the algorithm. First, it appears that the optimal factors 
$q(M)$, $q(N)$, and $q(\gamma)$ necessarily factorise as: 
$$
q(M) =\prod_{i=1}^{m_1}q(M_{i,\cdot}),\quad
q(N) =\prod_{j=1}^{m_2}q(N_{j,\cdot}),\quad
q(\gamma)  = \prod_{k=1}^K q(\gamma_k)
$$
where $q(M_{i,\cdot})$ is $\mathcal{N}(\mathbf{m}_{i,\cdot}^T,\mathcal{V}_i)$,
$q(N_{j,\cdot})$ is $\mathcal{N}(\mathbf{n}_{j,\cdot}^T,\mathcal{W}_j)$ and
$q(\gamma_k)$ is $\mathcal{IG}(a+(m_1+m_2)/2,b_k)$ for some $m_1\times K$ matrix $\mathbf{m}$
whose rows are denoted by $\mathbf{m}_{i,\cdot}$ and some $m_2\times K$ matrix $\mathbf{n}$
whose rows are denoted by $\mathbf{n}_{j,\cdot}$ and some vector $b=(b_1,\dots,b_K)$.
The parameters are updated iteratively through the formulas
\begin{enumerate}
 \item moments of $M$:
 $$\mathbf{m}_{i,\cdot}^T :=\frac{2\lambda}{n} \mathcal{V}_{i}
\sum_{k:i_k=i} Y_{i_k,j_k} \mathbf{n}_{j_k,\cdot}^T$$
$$
 \mathcal{V}_i^{-1} := \frac{2\lambda}{n} \sum_{k:i_k=i}
  \left[\mathcal{W}_{j_k} + \mathbf{n}_{j_k,\cdot} \mathbf{n}_{j_k,\cdot}^T \right]
    + \left(a+\frac{m_1+m_2}{2}\right)
    \mathbf{diag}(b)^{-1}$$
 \item moments of $N$:
 $$\mathbf{n}_{j,\cdot}^T :=\frac{2\lambda}{n} \mathcal{W}_{j}
\sum_{k:j_k=j} Y_{i_k,j_k} \mathbf{m}_{i_k,\cdot}^T $$
$$
 \mathcal{W}_j^{-1} := \frac{2\lambda}{n} \sum_{k:j_k=j}
  \left[\mathcal{V}_{i_k} + \mathbf{m}_{i_k,\cdot} \mathbf{m}_{i_k,\cdot}^T \right]
    + \left(a+\frac{m_1+m_2}{2}\right)
    \mathbf{diag}(b)^{-1}$$
 \item moments of $\gamma$:
 $$b_k:= \frac{1}{2}\left[
 \sum_{i=1}^{m_1}\left(\mathbf{m}_{i,k}^2 + (\mathcal{V}_i)_{k,k} \right)
 +
 \sum_{j=1}^{m_2}\left(\mathbf{n}_{j,k}^2 + (\mathcal{V}_j)_{k,k} \right)
 \right]$$
\end{enumerate}
(where
$(\mathcal{V}_i)_{k,k}$ denotes the $(k,k)$-th entry of the matrix $\mathcal{V}_i$
and
$(\mathcal{W}_j)_{k,k}$ denotes the $(k,k)$-th entry of the matrix $\mathcal{W}_j$).

\section{Link with minimization problems}
\label{section_MAP}

In this section, we highlight some connections between Bayesian estimation
based on a certain prior, as discussed in Section~\ref{section_priors}, 
and penalized estimators based on penalty terms that are popular for matrix completion (or other problems). More precisely, we show that, for a given prior, the 
MAP (maximum a posteriori), that is the mode of the posterior density, may 
be recovered as a certain penalized estimator. 
The motivation is to provide additional insight into the choice of  the prior distribution. In particular, we shall see that the gamma prior corresponds 
to a certain penalty function, which is popular and easy to interpret, but 
that may not be easy to implement directly. 

\subsection{Prior $\delta_1$}

When the prior is $\delta_1$, the MAP is
\begin{align*}
& \arg\min_{\theta = MN^T} \left\{
\frac{\lambda}{n}\sum_{k=1}^{n} (Y_k - \theta_{i_k,j_k})^2
+ \frac{\|M\|_F^2 + \|N\|_F^2}{2}
\right\} \\
& = \arg\min_{\theta = MN^T} \left\{
\frac{\lambda}{n}\sum_{h=1}^{n} (Y_h - \theta_{i_h,j_h})^2
+ \sum_{\ell=1}^K \frac{\|M_{\cdot,\ell}\|^2 + \|N_{\cdot,\ell}\|^2}{2}
\right\}.
\end{align*}
The penalization is very similar to the ridge penalty used for regression
problems.
It is a classical result that (when $K=m_1\wedge m_2$),
$$ \| \theta\|_{*} =  \inf_{M,N: \theta=MN^T} \frac{\|M\|_F^2 + \|N\|_F^2}{2} $$
where $\|\theta\|_*$ is the nuclear norm of $\theta$ (see {\it eg} Equation 2
page 203
in~\cite{Recht-Re} or Lemma 1 in~\cite{Srebro1}). So, this MAP can be rewritten
as
$$
\arg\min_{\theta} \left\{
\frac{\lambda}{n}\sum_{h=1}^{n} (Y_h - \theta_{i_h,j_h})^2
+ \|\theta\|_*
\right\}
$$
and linked with the penalization problems studied
in~\cite{Recht-Re,Rohde,Klopp,KLT}.

\subsection{Inverse-gamma prior}

When we use the prior $\Gamma^{-1}(a,b)$, the MAP is
\begin{multline*}
\arg\min_{\theta = MN^T,\gamma} \Biggl\{
\frac{\lambda}{n}\sum_{k=1}^{n} (Y_k - \theta_{i_k,j_k})^2
\\
+ \frac{1}{2}\sum_{\ell=1}^K \left[ \frac{\|M_{\cdot,\ell}\|^2
   + \|N_{\cdot,\ell}\|^2 + b}{\gamma_{\ell}} + (a+1)\log (\gamma_\ell) \right]
\Biggr\}.
\end{multline*}

When the $\gamma_h$ are fixed, we can interpret this as a weighted ridge
regression.
For small $\gamma_h$, $M_{\cdot,\ell}$ and $N_{\cdot,\ell}$ are close to zero.
So, the
essential rank of $\theta$ is the number of $\gamma_h$ that are not too small.
On the
other hand, the penalization $(a+1)\log (\gamma_h)$ will cause many $\gamma_h$
to be
small.

\subsection{Gamma prior}

When we use a prior $\Gamma(a,b)$, the MAP is
\begin{multline*}
\arg\min_{\theta = MN^T,\gamma} \Biggl\{
\frac{\lambda}{n}\sum_{k=1}^{n} (Y_k - \theta_{i_k,j_k})^2
\\
+ \frac{1}{2}\sum_{\ell=1}^K \left[ \frac{\|M_{\cdot,\ell}\|^2
   + \|N_{\cdot,\ell}\|^2 }{\gamma_{\ell}} - (a-1)\log (\gamma_\ell) +
b\gamma_\ell \right]
\Biggr\}.
\end{multline*}

The interpretation is similar to the one in the inverse gamma case.
Note that, we used this prior with parameters $(a,b)=(m_1+m_2+1)/2,\beta^2 /2)$. In this
case,
there is an interesting phenomenon. If we do not consider the
MAP with respect to $M$, $N$ and $\gamma$, but instead integrate with respect to
$\gamma$
and only consider the MAP with respect to $M$ and $N$, the estimator is actually
similar to Yuan and Lin's group-LASSO estimator~\cite{GLASSO}.

\begin{prop}
\label{prop_glasso}
 The MAP of the marginal posterior distribution of $\theta$ under the Gamma
prior
 is given by
$$
\arg\min_{\theta = MN^T} \Biggl\{
\frac{\lambda}{n}\sum_{k=1}^{n} (Y_k - \theta_{i_k,j_k})^2
+ \beta \sum_{\ell=1}^{K} \sqrt{ \|M_{\cdot,\ell}\|^2 +  \|N_{\cdot,\ell}\|^2 }
\Biggr\}.
$$
\end{prop}

The proof is given in Section~\ref{section_proofs}. Note that, contrary
to
the group-LASSO, this optimization problem has no reason to be convex, and might
not lead to feasible algorithms for large scale problems. 

On the other hand, it gives a nice extra motivation for the gamma prior: 
this prior tends to set some columns of $M$ and $N$ to
$0$
in the same way than the group LASSO set some group of coefficients to $0$
simultaneously.

\section{Simulations}
\label{section_simulations}

We first compare all the Bayesian estimators corresponding to the different
priors on a toy example. Following~\cite{Candes1}, we generate a square matrix
$\theta^0$ ({\it ie} $m_1=m_2=m$)
with rank $r = 2$ in the following way: $\theta^0 = M^0 (N^0)^T$ where the entries
of the $m\times 2$ matrices $M^0$ and $N^0$ are iid $\mathcal{N}(0,20/\sqrt{m})$.

We observe $20\%$ of the entries matrix, $n=0.2 m^2 $, corrupted by a $\mathcal{N}(0,1)$ noise. The performance of each estimator $\hat{\theta}$
is measured through its ${\rm RMSE}$
$$ {\rm RMSE} = \sqrt{\frac{1}{m_1 m_2}\|\hat{\theta}-\theta^0\|_F^2}
   = \frac{1}{m} \|\hat{\theta}-\theta^0\|_F .$$

In a first set of experiments, we fix $K=5$ and study the convergence of
the estimators when $m$ grows, $m\in\{100,200,500,1000\}$. In a second set of experiments,
we fix $m=500$ and study the effect of $K$ on the performance of the different
estimators, $K\in\{2,5,10,20\}$.

\begin{figure}
\includegraphics[scale=.6]{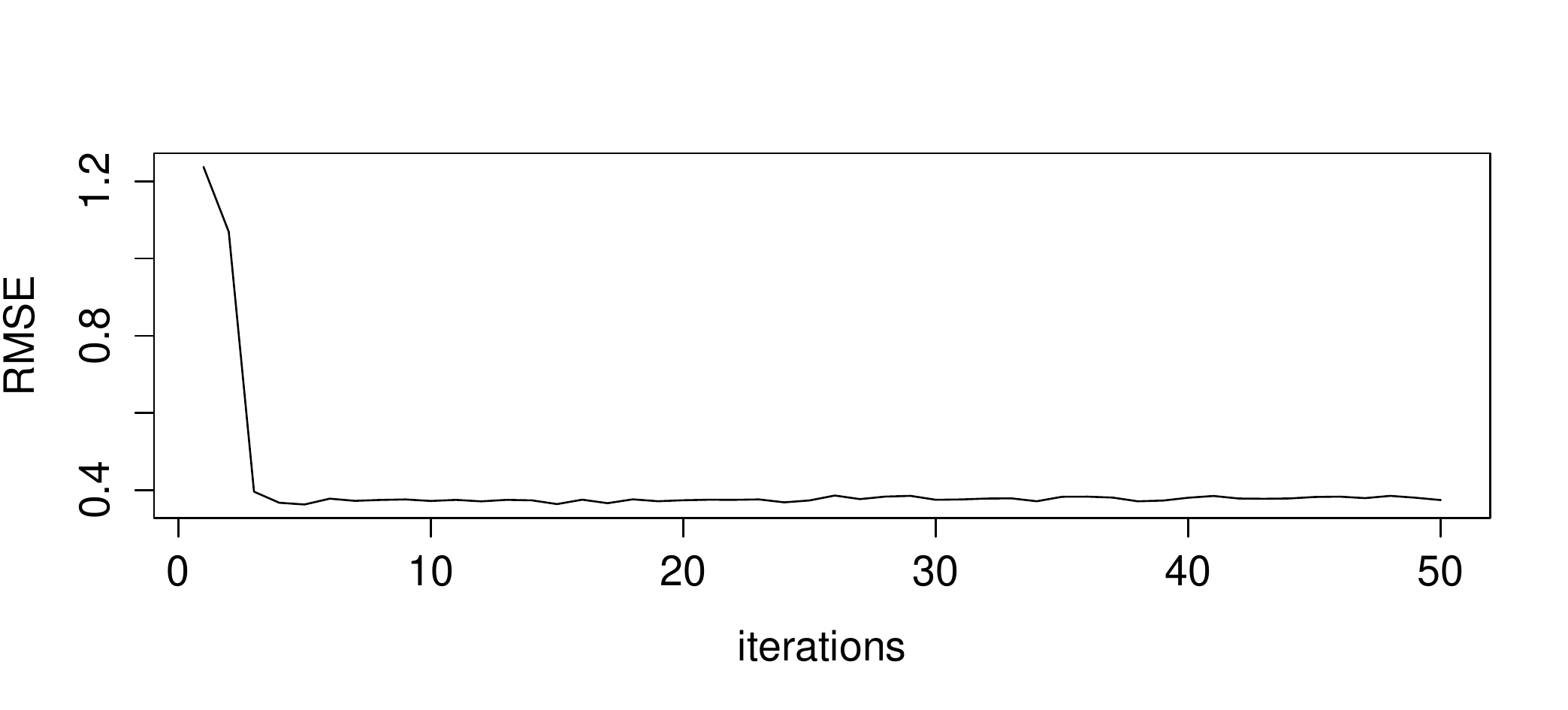}
\caption{RMSE by iterations, s for $m=200$, $K=5$
with the Bernoulli prior with $(C,p,\varepsilon)=(1,0.05,0.05)$.}
\label{RMSEplot}
\end{figure}

\begin{figure}
\includegraphics[scale=.6]{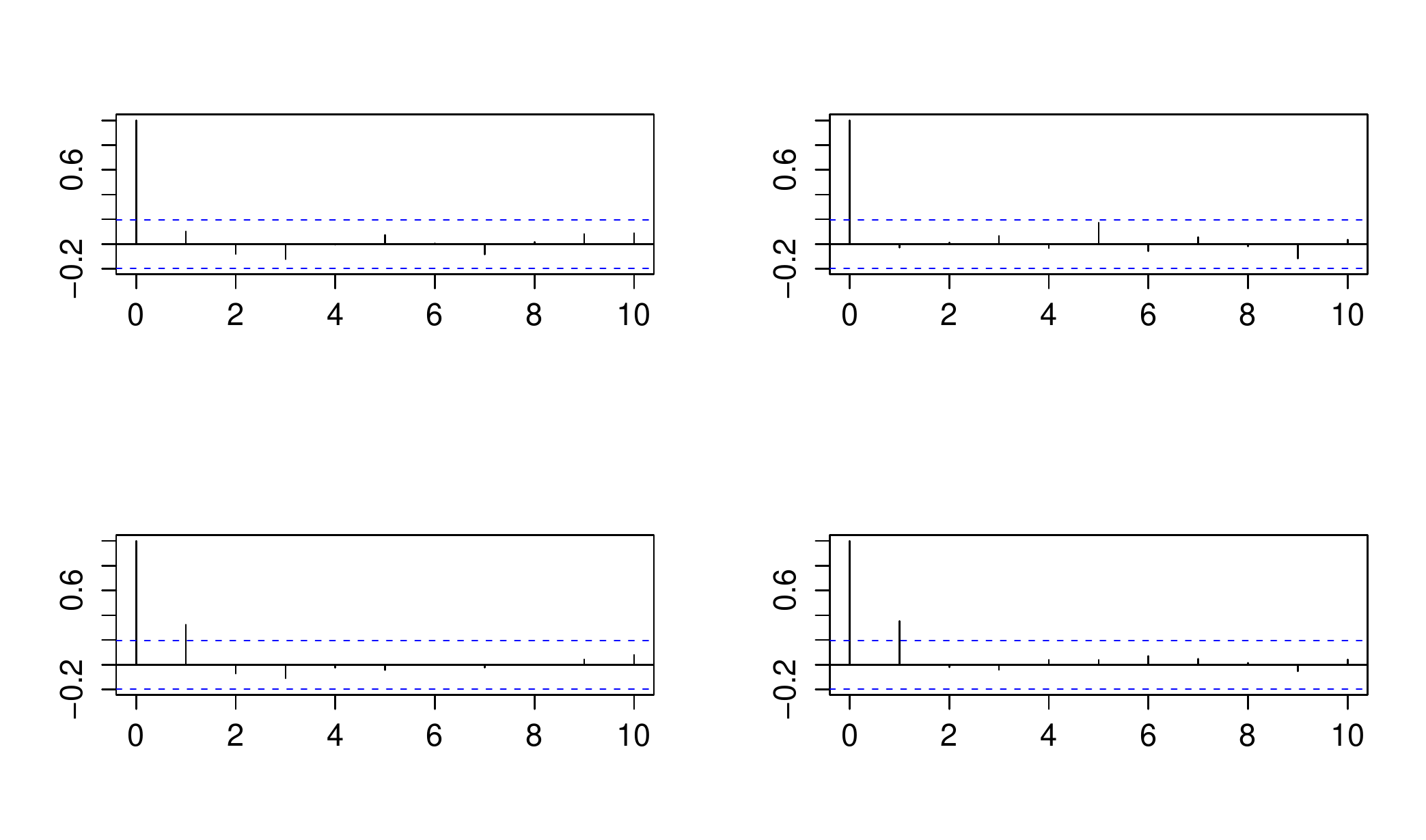}
\caption{ACF for some entries of $\theta$, s for $m=200$, $K=5$
with the Bernoulli prior with $(C,p,\varepsilon)=(1,0.05,0.05)$.}
\label{ACFplot}
\end{figure}

Note that in this simulation study, we always use Gibbs sampling to simulate
from the posterior. The convergence of the chain seems very quick, as illustrated
by Figure~\ref{RMSEplot}, which is taken from one of the simulations for $m=200$, $K=5$
with the Bernoulli prior with $(C,p,\varepsilon)=(1,0.05,0.05)$.
The autocorrelations 
for each entry of the matrix $\theta$ are not large and vanishes after 2 or 3 lags, as shown by Figure~\ref{ACFplot}.
In any case, we let the Gibbs sampler run $1000$ iterations and remove the first $100$
iterations as a burn-in period. The thinning parameter is set to $10$.

\subsection{Convergence when $m$ grows}

The results of the experiments with $K$ fixed are reported in Table~\ref{tab:mgrows}.
\begin{table}
\caption{Experiments with fixed $K$: RMSE for different priors and different values of $m$.}
\label{tab:mgrows}
\begin{center}
\begin{tabular}{lcccc}
& \multicolumn{4}{c}{$m$} \\
prior distribution		& $100$	& $200$	& $500$	& $1000$  \\
Fixed			& 	.75	&	.47	&	.27	&
.18	\\
Gamma			&	.60	&	.37	&	.23	&
.16	\\
Inverse Gamma 	&	.59	&	.39	&	.25	&   .18	\\
Discrete		&  .60	&	.36	&	.22	&	.16	
\end{tabular}
\end{center}
\end{table}
For each prior, we tried many hyperparameters and only report the best results.
The corresponding hyperparameters are reported in Table~\ref{tab:mgrows2}.
\begin{table}
\caption{Experiments with fixed $K$: choice of hyperparameters for the different
considered priors.}
\label{tab:mgrows2}
\begin{center}
\begin{tabular}{lcccc}
& \multicolumn{4}{c}{$m$} \\
prior distribution		& $100$	& $200$	& $500$	& $1000$  \\
Fixed			& $\gamma=0.2$ & $\gamma = 1$ & $\gamma=7$ & $\gamma=10$
\\
Gamma			& $ \beta^2=500$&$ \beta^2=2000$&$ \beta^2=10000$&$ \beta^2=40000$\\
Inverse Gamma ($a=1$)	&$b=0.015$&$b=0.012$&$b=0.005$&$b=1,0.007$\\
Discrete ($(C,p)=(1,0.05)$)		&$\varepsilon=0.11$
  &$\varepsilon=0.08$&$\varepsilon=0.05$&
   $\varepsilon=0.03$
\end{tabular}
\end{center}
\end{table}
First, it appears that the results from the four different priors are very close.
Note that the choice $a=1$ and $b\ll 1$ in the inverse Gamma distribution was done according
to the theoretical results in~\cite{Alquier} in reduced rank regression.
The non-adaptive prior always performs worse than the adaptive priors though,
as expected in this case where $r=2<K=5$.

The results improve when $m$ grows, in line with a possible consistency of
the estimator when $n$, $m_1$ and $m_2$ grow.

\subsection{$m=500$}  

We focus on the case $m=500$ to explore the behaviour of the results when the
size of the parameters varies. The rationale beyond this is that, in real-life
applications, we don't know the rank $r$ of $\theta^0$, so we usually set
$K$ ``too large'' and hope for adaptation as explained in Section~\ref{section_priors}.
The results are reported in Table~\ref{tab:Kgrows}.
\begin{table}
\caption{Experiments with fixed $m$: RMSE for different priors and different choices for $K$.}
\label{tab:Kgrows}
\begin{center}
 \begin{tabular}{lcccc}
    	& \multicolumn{4}{c}{prior distribution} \\
	  & Fixed   & Gamma 	& Inv. Gamma    	& Discrete \\
$K=2$ &  .22	&		.22	&	.22		&.22	
\\
$K=5$ &	 .27   	&		.23	&	.25		&.22	
\\
$K=10$&	 .31   	&		.23	&	.26		&.22	
\\
$K=20$&	 .37   	&		.22	&	.27		&.22
\end{tabular}
\end{center}
\end{table}
The corresponding hyperparameters are reported in Table~\ref{tab:Kgrows2}.
\begin{table}
\caption{Experiments with fixed $m$: choice of hyperparameters for the
different considered priors.}
\label{tab:Kgrows2}
\begin{center}
 \begin{tabular}{lcccc}
    	& \multicolumn{4}{c}{prior distribution} \\
	  & Fixed   & Gamma 	& Inv. Gamma, $a=1$& Discrete, $(C,p)=(1,0.05)$ \\
$K=2$ &$\gamma=1$&$\beta^2=5000$&$b=0.001$&$\varepsilon=0.05$
\\
$K=5$ &$\gamma=7$&$\beta^2=10000$&$b=0.005$&$\varepsilon=0.05$	
\\
$K=10$&$\gamma=6$&$\beta^2=12500$&$b=0.006$&$\varepsilon=0.03$
\\
$K=20$&$\gamma=6$&$\beta^2=13000$&$b=0.003$&$\varepsilon=0.02$
\end{tabular}
\end{center}
\end{table}
As expected, the non-adaptive $\delta_1$ prior does not lead to stable results and
get worse when $K$ is too large. The lowest RMSE is achieved when the size is equal
to the true rank.
The three ``adaptive'' estimators perform better. However, it is to be noted
that the best performance is reached by the gamma and the discrete distributions,
while the inverse gamma is the most popular in the literature. As expected, these
priors adapt automatically to the true rank of the matrix $\theta^0$, $r=2$, and
taking a large $K$ do not deteriorate the performances of the estimators.

Note that the (rather) poor performance of the inverse gamma distribution seems to be
caused by a slower convergence of the MCMC algorithm in this context, and/or by
the heavy-tails of the inverse gamma distribution.

\section{Test on MovieLens}
\label{section_realdata}

We now test the Bayesian estimators with the discrete prior and with the inverse
gamma prior on the MovieLens dataset, available online:

\noindent {\it http://grouplens.org/datasets/movielens/}

There are actually three different datasets with respectively about 100K, 1M and 10M
ratings. Note that this is a challenging situation, because the size of the matrix
$\theta^0$ makes the Gibbs iteration very slow, preventing from doing $1000$ iterations,
at least in the 1M and 10M cases. In this case, it is tempting to use the Variational
Bayes algorithm (VB)~\cite{Bishop6} instead of the Gibbs sampler. VB has been used on the
Netflix challenge in~\cite{Lim} with a similar model.

The dataset is split into two parts, the training set (80 \%) and the test set
(20 \%). The model is fitted on the training set and we measure the RMSE on the
other part.

\begin{figure}
\center
\includegraphics[scale=.7]{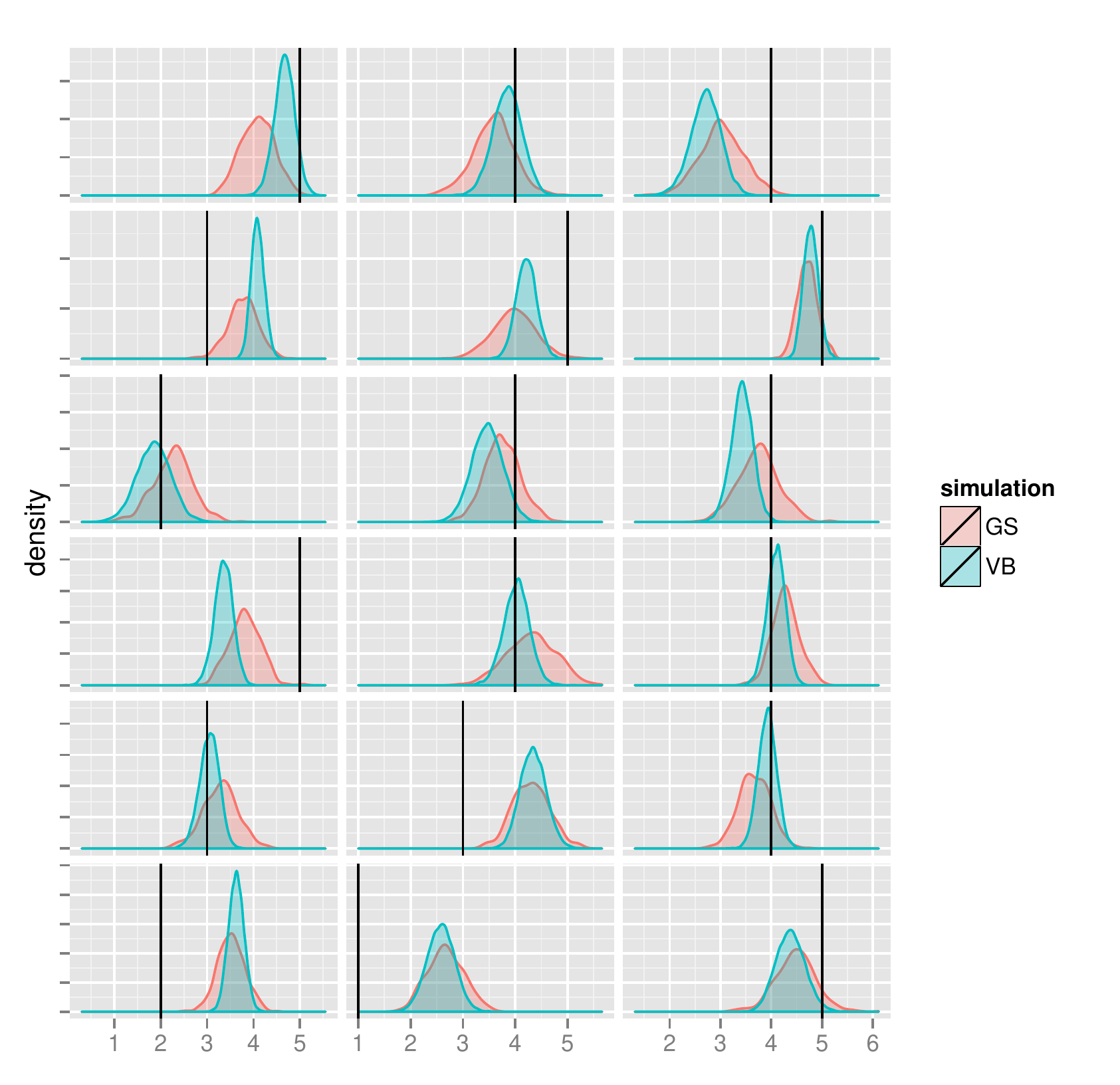}
\label{compare_GS_VB}
\caption{Posterior distribution approximated by Variational Bayes (VB) and the
Gibbs sampler (GS) on a few randomly selected
entries of the matrix. The prior is the inverse gamma prior.}
\end{figure}

\begin{figure}
\center
\includegraphics[scale=.4]{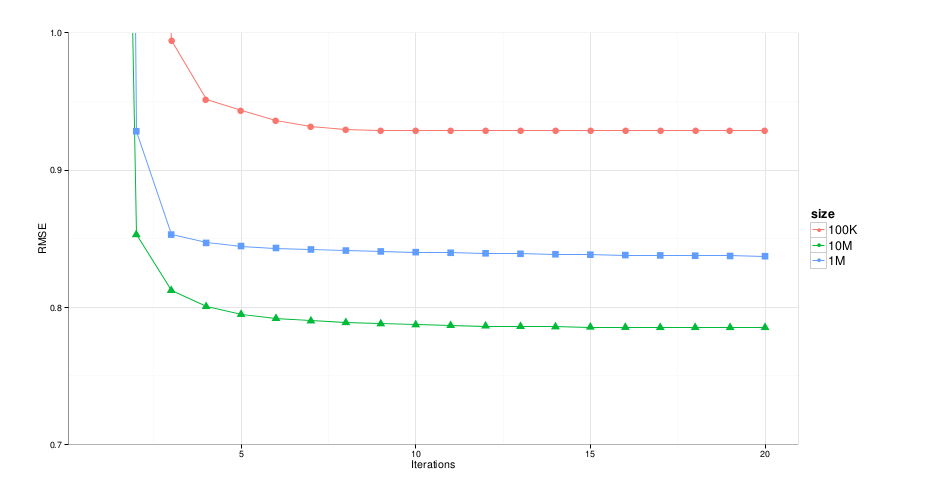}
\caption{Convergence of the VB algorithm on the three MovieLens dataset.}
\label{figureCV}
\end{figure}

First, we compared VB and Gibbs on the 100K dataset in Figure~\ref{compare_GS_VB}.
It appears that the distribution of the matrix is quite different at least for a
few entries, but in the end, the performance of the approximation in terms of RMSE
are comparable, as shown by Table~\ref{tab:movielens}.
\begin{table}
\caption{Tests on the various MovieLens datasets.}
\label{tab:movielens}
\begin{center}
 \begin{tabular}{lcccc}
 Dataset 	& Algorithm 		& prior & hyperparameters & RMSE \\
 \hline
 100K 		& GS	& Discrete	& $(C,p,\varepsilon)=(1,0.05,0.07)$
 &  .92  \\
 100K 		& GS	& Inverse Gamma	& $(a,b)=(1,0.1)$ &  .92  \\ 
 100K 		& VB & Inverse Gamma	& $(a,b)=(1,0.1)$  &  .92 \\
 \hline
 1M 		& VB & Inverse Gamma	& $(a,b)=(1,0.1)$  &  .84 \\
 10M 		& VB & Inverse Gamma	& $(a,b)=(1,0.1)$  &  .79  
\end{tabular}
\end{center}
\end{table}
Also, in this case, the estimator based on the inverse gamma prior performs
as well as the one base on the discrete prior. So, we only used the inverse
gamma prior with the VB algorithm for the more time-consuming tests on the 1M
and 10M dataset. As shown by figure~\ref{figureCV}, even in the 10M dataset,
the VB algorithm converges in less than 20 iterations.

\section{Conclusion}

We reviewed the popular priors in Bayesian matrix completion (non-adaptive and
inverse gamma priors) and proposed two new priors (gamma and discrete priors).
We demonstrated the efficiency of all these estimators on simulated and real-life
datasets. Future work should include the study of the optimality of these estimators.
Extension to tensors
in the spirit of~\cite{recht-tensors,suzuki-tensors}, would also be of interest.

\section{Proofs}
\label{section_proofs}

\noindent {\it Proof of Proposition~\ref{prop_glasso}:}
This idea of the proof comes from a similar argument for the Bayesian
LASSO in~\cite{Park-Casella}, namely, to use the formula
\begin{equation}
 \label{formula-park-casella}
\int_{0}^{\infty} \frac{1}{\sqrt{2\pi s}} \exp\left(
\frac{z^2}{2s}
\right) \frac{a^2}{2} \exp\left(-\frac{a^2 s}{2}\right) {\rm d}s = \frac{a }{2}
 \exp\left(-a|z|\right)
\end{equation}
for any $a$ and $z$.
We have:
\begin{align*}
 \rho_{\lambda}(M,N)
  & = \int \rho_{\lambda}(M,N,{\rm d}\gamma) \\
  & = \int \exp\left[ - \frac{\lambda}{n} \sum_{k=1}^n(Y_{k} -
(MN^T)_{i_k,j_k})^2  \right]
\pi(M,N,{\rm d} \gamma) \\
  & = \exp\left[ - \frac{\lambda}{n} \sum_{k=1}^n(Y_{k} - (MN^T)_{i_k,j_k})^2 
\right]
  \int \pi(M,N,{\rm d} \gamma)
\end{align*}
and then,
\begin{align*}
  \int \pi(M,N,{\rm d} \gamma)
   & \propto \prod_{\ell=1}^{K} \int_{0}^{\infty}
\gamma_\ell^{-\frac{m_1+m_2}{2}}
   \exp\left[-\frac{1}{2\gamma_k}\left(\|M_{\cdot,k}\|^2 +
\|N_{\cdot,k}\|^2\right) \right]
    \\ & \quad \quad \quad \quad \quad \quad 
   \gamma_\ell^{\frac{m_1+m_2+1}{2}} \exp\left(-\frac{\beta^2}{2} \gamma_{\ell}
\right)
    {\rm d}\gamma_{\ell}
   \\
   & \propto \prod_{\ell=1}^{K} \int_{0}^{\infty} \gamma_\ell^{\frac{1}{2}}
   \exp\left[-\frac{1}{2\gamma_k}\left(\|M_{\cdot,k}\|^2 +
\|N_{\cdot,k}\|^2\right)
   -\frac{\beta^2}{2} \gamma_{\ell}\right]{\rm d}\gamma_{\ell}
   \\
   & \propto \exp\left[
   -\beta \sum_{\ell=1}^{K} \sqrt{\|M_{\cdot,k}\|^2 + \|N_{\cdot,k}\|^2}
   \right]
\end{align*}
using~\eqref{formula-park-casella}. This ends the proof. $\square$

\section*{Acknowledgements}

We would like to thank Prof. Taiji Suzuki for insightful comments.

\bibliographystyle{alpha}
\bibliography{biblio}

\end{document}